\documentclass[]{ceurart}

\usepackage{listings}
\usepackage{booktabs}

\copyrightyear{xx}
\copyrightclause{}

\conference{VIPERC2025: 4th International Conference on Visual Pattern Extraction and Recognition for Cultural Heritage Understanding, 3-4 December 2025}
\copyrightyear{2025}
\copyrightclause{Copyright for this paper by its authors. Use permitted under Creative Commons License Attribution 4.0 International (CC BY 4.0).}

\begin{document}

\title{Empowering Clients: Transformation of Design Processes Due to Generative AI}

\author[1]{Johannes Schneider}[email=johannes.schneider@uni.li]
\cormark[1]
\author[1]{Sinem Kilic}[email=kilic.sinem@uni.li]
\author[1]{Daniel Stockhammer}[email=daniel.stockhammer@uni.li]
\address[1]{University of Liechtenstein, Vaduz, Liechtenstein}

\cortext[1]{Corresponding author.}

\begin{abstract}
Generative AI (GenAI) is transforming creative fields shaping our culture and our heritage. We focus on wide-spread interactions between clients and (creative) specialists highlighting a change in interaction patterns leading to a shift from the use of expert creativity towards AI-supported client creativity. More specifically, we explore the case of architecture as designing houses is complex involving extensive customer interaction. We investigate the effects of GenAI on the architectural design process and discuss the role of the architect. Our study involved six architects using a general-purpose text-to-image tool for generating designs and providing feedback followed by expert interviews. We find that AI can disrupt the ideation phase by enabling clients to engage in the design process through rapid visualization of their ideas. In turn, so our thesis, the architect’s role shifts towards assessing feasibility of such designs. AI’s feedback, though valuable, can hamper creativity and innovation by suggesting altering novel, innovative approaches towards more standardized designs. We find that there is uncertainty among architects about the interpretative sovereignty of architecture and identity when AI increasingly takes over authorship. Our findings can also support the design of future AI systems by pinpointing weaknesses and highlighting a novel design process calling for tighter client integration. In our discussion, we also generalize our findings on a broader societal level elaborating on the change of a number of characteristics such as power, capability and responsibility in the triangle of AI, experts, and non-experts. We also discuss risks such as cultural uniformity when it comes to using AI to design artifacts central to our cultural heritage. 
\end{abstract}


\maketitle

\section{Introduction}
Societal change driven by Artificial Intelligence (AI) is reshaping professional roles, altering creative processes, and redefining traditional workflows across various domains \cite{r1}. Many fields associated with creativity such as architecture are undergoing a profound transformation driven by the rise of AI, especially Generative AI (GenAI) \cite{r2}. GenAI promises unprecedented potential for innovation and efficiency in creative design processes. At the same time it also poses a threat of machines monopolizing creativity \cite{r3} leading potentially to cultural uniformity. Specifically, text-to-image models, such as Dall-E, have emerged as valuable tools, enabling creative professions such as architects and visual artists, but also laymen, to envision, iterate, and communicate designs in ways previously constrained by traditional methodologies requiring extensive expertise \cite{r4}. These models challenge the boundaries of traditional architectural practice. They transform, at best, enhance the ideation phase and extend the architect’s “cognitive workbench” beyond conventional approaches \cite{r5, r6}. But they are also being met with aversion \cite{r7}.

Prior work has focused primarily on the supporting role of GenAI for experts in ideation, e.g., \cite{r8}, or comparing different tools for ideation subtasks, e.g., \cite{r9}. In contrast, we consider the architectural design process more holistically. Our work also the clients' perspective and using AI not only for the most creative part — ideation — but also for feedback, which has received relatively little attention, e.g., \cite{r10}. More broadly, our work touches on the unexplored notion of augmentation vs.\ automation within the architectural profession \cite{r11} as well as computer-mediated creativity in design processes involving also customers \cite{r12}. At the same time, we also elaborate on wider societal impacts including the cultural heritage that might be strongly influenced by AI.

The rise of GenAI naturally raises new questions regarding the architect's role and authorship due to the implementation of AI in architecture. Thus, our research questions are:

\noindent\textbf{RQ 1:} How can GenAI support (or fail to support) the design process including client engagement?

\noindent\textbf{RQ 2:} How is the role of architects and its clients changing?

In this work we provide first answers based on letting architects solve a task using GenAI followed by interviews involving six architects. The study reveals that GenAI can empower customers and architects alike to create fast design visualizations making existing process steps such as sketching designs less relevant. However, as of today, AI comes with strong limitations creating potentially credible visualizations that are infeasible to build or disregard important contextual knowledge, e.g., characteristics of a plot, structural or building physics requirements. Thus, false expectations might emerge, and the architects must carefully assess designs. Interestingly, tasks often considered less creative such as turning designs into detailed plans — financially the most lucrative phase of work for architects — have been less impacted by GenAI.

In turn, architects in our study express concern over the potential erosion of their creative agency and the profession’s social perception. In our discussion, we illustrate the changed design process focusing on the interaction of the creative professional, clients, AI and the design artifact. In addition, we discuss more broadly implications on a number of dimensions such as culture, power, capability, and responsibility for experts, non-experts, and AI.

\section{Related Work and Background}
Long before GenAI creativity has been studied in fields such as computational creativity, e.g., \cite{r15}. The transformative nature of GenAI meaning the rise of large multi-modal foundation models \cite{r16} showing remarkable creative capabilities has dramatically accelerated the discussion on GenAI’s impact on work. GenAI has been shown to successfully augment creative work \cite{r17}. In the realm of urban planning, the benefits of GenAI have been demonstrated for designing blue prints (defining the shape of houses seen from above) \cite{r18}. The investigated AI system did not allow for customization, i.e., prompting, but rather it just generated proposals. Our findings are complementary to Weber’s as there is little overlap, which is not surprising as we focus on a different task, i.e., designing a building including details rather than just the outline, and we used advanced text-to-image models which are anticipated to alter the architectural visual culture \cite{r19, r20}. These models support the creative thinking process \cite{r21} but are also viewed critically. \cite{r22} argues that the generation of images can be very time consuming, as the desired results can only be achieved through prompt engineering driven by trial and error. \cite{r23} adds that text-to-image models such as Stable Diffusion, Midjourney, and Dall-E are yet unable to incorporate specific styles in design and shapes in the generated images though they can help, i.e., to turn design sketches into 3d designs \cite{r9}. \cite{r8} conducted a laboratory study involving students only and various text-to-image tools showing that GenAI can be a helpful tool for architects for ideation, but also highlight limitations such as the inability to generate floor plans. Our findings confirm technical shortcomings through statements from professionals. Our focus is on professionals and the entire design process covering additional use-cases of GenAI such as feedback and stakeholders revealing further shortcomings more relevant in practice. A number of works also aimed to design special AI models, e.g., for exterior conceptual design \cite{r24} and architectural visualizations \cite{r25}. Such models and ideas might be accessible in the future for architects, but many clients of architects are more likely to use general purpose AI of commercial AI vendors. The weaknesses outlined in general-purpose AI might also inform future studies to build better tailored tools for both specialists and laypeople. 

Design is an intricate, iterative process with focus on problem-solving activities encompassing diverse cognitive skills, such as the intangible elements of intuition, imagination, and creativity \cite{r26}, \cite{r27}. Within the design process, a set of configurations are proposed and reviewed to achieve intended design purposes. However, design is intended to transcend mere appropriateness and functionality, encompassing the inception and materialization of novel ideas, effectively balancing intellectual pursuits with technical considerations \cite{r28}. Architectural design can be approached as a problem-solving practice for which architects design human-centric and functional spaces and environments to accommodate diverse human activities \cite{r29}. By necessitating that architectural creations simultaneously possess a blend of aesthetic appeal, structural integrity, and practical functionality, the discipline of architecture sets itself apart from other (purely) artistic disciplines \cite{r30}. Architects are required to possess crucial cognitive skills including creativity and spatial awareness \cite{r26}.

\section{Methodology}
In our evaluation we ask experts, i.e., architects, to use a text-to-image model followed by semi-structured interviews. The hands-on evaluation ensures that all participants get exposure to GenAI, in particular the evaluated tool Dall-E 3, directly prior to the interviews. This ensures informed perspectives grounded in actual interaction and a consistent baseline as not all architects were familiar with GenAI tools. By using an offline setting (in contrast to an online setting, observing architects in a real-world project), we facilitate interpretation of outcomes \cite{r31}. We chose Dall-E 3 as a tool  due to its wide accessibility and usability, i.e., it is integrated within ChatGPT. It can be considered as an easy to access state-of-the-art tool for image generation that can be used by architects and their clients, which is important as we seek to understand GenAI usage holistically including also the potential usage by laymen for tasks so far executed by experts. For the creation of the experiment, the process outlined by \cite{r31} was used. While it is more common in a quantitative setting in software engineering, we found it valuable for the experiment design as it provides a structured step-by-step approach to experimentation and evaluation of software, in particular, we leveraged the scoping, planning and operation phase. We set the scope to a single object, i.e., a single GenAI tool, and multiple subjects (i.e., architects) as our primary concern is understanding people’s views and most popular general-purpose GenAI tools such as Midjourney, Dall-E and stable diffusion perform qualitatively similar.

The hands-on session lasted 60 min including 10 min introduction. The task was to construct a single-family home or hotel. Participants were asked to follow the classic design process of ideation, concept generation and design visualization and employ GenAI for feedback generation. For the semi-structured interviews, we asked questions derived from RQ 1 and RQ 2.

The expert interviews were transcribed, analyzed, and embedded into the context of the research topic \cite{r32}. To this end, \cite{r33} proposed applying the so-called hermeneutic approach, which enables the comparison of opinions across different interviews to “try to make sense of the whole, and the relationship between people, the organization, and information technology.“ In the first two cycles of interpretation one author transcribed and coded the interviews, followed by a second author who went through the interviews and assessed the coding, aligning with the first (and other authors) in case of disagreements. As these authors were potentially biased towards AI, in a further round of interpretation, an architect complemented and assessed the interpretation discussing with the other authors, which only led to minor adjustments. The interview participants (Table~\ref{tab:interview-partners}) consisted of four architects with significant work experience and two students.

\begin{table}[ht]
\centering
\caption{Interview partners}
\label{tab:interview-partners}
\scriptsize
\begin{tabular}{lccccccc}

\textbf{Int.} & \textbf{Age} & \textbf{Gen.} & \textbf{Ed.} & \textbf{Study Years} & \textbf{Work Exp.} & \textbf{Exp. With AI} & \textbf{Time} \\
IP1 & 24 & Male & Master & 6 & 3.5 & No & 40 \\
IP2 & 28 & Male & Master & 5 & 5 & Yes & 35 \\
IP3 & 22 & Female & Bachelor & 3 & 0.5 & No & 30 \\
IP4 & 22 & Male & Bachelor & 3.5 & 0 & No & 32 \\
IP5 & 32 & Female & Master & 6 & 8 & Yes & 28 \\
IP6 & 42 & Female & Master & 5 & 16 & No & 27 
\end{tabular}
\end{table}

\section{Results}
\begin{figure*}[ht]
\centering
\includegraphics[width=0.7\linewidth]{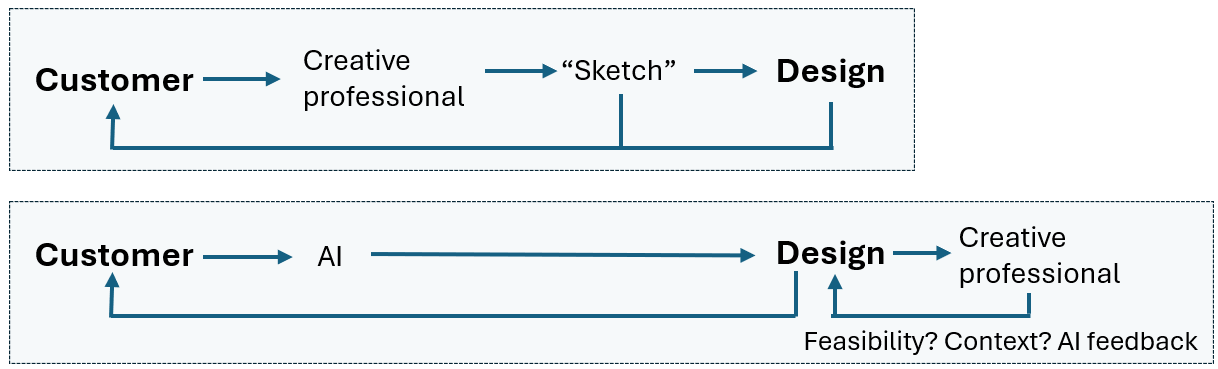}
\vspace{-3pt}
\caption{Abstracted design process without AI (top) and with AI used by clients (bottom).} \label{fig:process}
\vspace{-6pt}
\end{figure*}

We first discuss RQ 1 discussing the impact on ideation and design visualization as well as client interaction. Figure~\ref{fig:process} summarizes key findings visually focusing on a design process driven conjointly by clients and GenAI. We also discuss the value of GenAI for architects.

\subsection{Ideation by architects}
All interview participants indicated that AI can serve as a source of inspiration and provide reference models during the ideation phase. IP4 emphasizes the value of GenAI for brainstorming and IP1 for generating ideas and exploring them further or in coming up with entirely new concepts. IP2 stated that AI is particularly useful when an architect lacks a clear vision of the project's final form. IP6 emphasized that the occassionally the proposed design as a whole was of limited value, but some design elements were inspirational. By inputting key project data into AI, the tool can take over the initial creative thinking process, offering the first potential ideas for implementation and making the concept more tangible for the architect.

\begin{figure}[ht]
\centering
\includegraphics[width=0.7\linewidth]{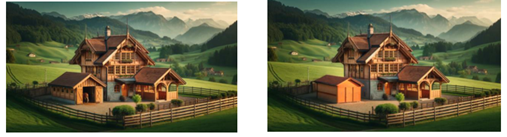}
\vspace{-3pt}
\caption{AI fails to follow the prompt ``The barn should be structurally connected to the house'' to modify the left panel (IP2), but it only closes the door (right panel).}\label{fig:prompt}
\vspace{-6pt}
\end{figure}

Compared to the aspects of human expertise and capability required for design development, IP5 and IP6 bring attention to the standardization and procedures in architectural work. While architecture is a creative field, it is also bound by numerous building regulations and rules. This duality creates a complex environment where creative design must align with stringent regulatory requirements. However, AI tools currently struggle to navigate this balance, often failing to integrate creative aspects with necessary building regulations. IP2 criticizes that AI often randomly assembles elements based on the inputs instead of reproducing human imagination in a visual form. IP1 says that AI paraphrases the given input information in a figurative sense. This can lead to a mismatch between AI-generated designs and their intended use (IP3). For instance, it is considered problematic when AI selects a single-family house design for a large complex building such as a hotel. Also designs are not extra-ordinary though appealing (IP6), hinting at AI’s creativity limits.
\paragraph{Prompt following — Inaccurate AI designs}
All interview partners argue that AI shows limited capability of meeting precise requirements which may in fact slow down the design process, as it often fails to generate exactly what is needed leading to rework as illustrated in Figure~\ref{fig:prompt}. IP4 states that AI requires extensive trial and error to achieve the desired results and notes that within the architectural field, there is little room for AI's visual interpretation as precise outputs are necessary based on the input specifications.

\begin{figure}[ht]
\centering
\includegraphics[width=0.7\linewidth]{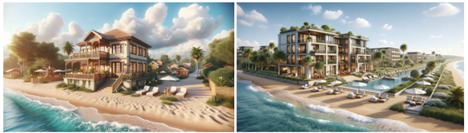}
\vspace{-3pt}
\caption{Incremental design issues: Asking for a hotel to be more luxurious (left) leads to a completely different design (right).}\label{fig:incremental}
\vspace{-6pt}
\end{figure}

IP6 critized AI for not altering its design although it was told about shortcoming with respect to regulations, while IP3 noted that designs can vary considerably with minor changes in instructions, e.g., asking for a hotel to be more luxurious caused a completely different buildings without similarities to the original one. Thus, an incremental design process, where most of the current elements are maintained and only parts are changed is difficult in practice (Figure~\ref{fig:incremental}).

\paragraph{Skipping sketching — The absence of imagination in visualizations}
IP4 notes that by using AI for the ideation phase, the sketching phase is skipped entirely as by inputting design ideas, the tool delivers final renderings instead of sketches. IP3 says that AI generated images can be considered as “rendering-sketches” meaning that they do not have the level of conception as a typical rendering (although appearing as such) but go beyond sketches in the elaboration. However, architects still see value in sketching. IP6 sees sketching as supporting her own ideation. When a client requests a building design, an architect often has a mental image that is not yet verbalized. This image is then quickly translated into a sketch with pen and paper, allowing the idea to be presented and explained to the client. Additionally, when modifications are needed, manual drawing is stated to be much faster as a piece of tracing paper can be placed over the draft to quickly sketch changes (IP3, IP6). Moreover, the feasibility issues in AI-generated architectural designs are addressed. An image created with AI may look realistic, but its practicality is questionable due to its form, position, supporting structure, and placement of individual objects as for example, the building might lack a connection to the street or a load-bearing column is not connected to the static system and foundation. As diverse as AI-generated designs may be, the default image without a specific prompt is a standard rendering that borders on kitsch, with little scope for personal interpretation and imagination.

\begin{figure}[ht]
\centering
\includegraphics[width=0.7\linewidth]{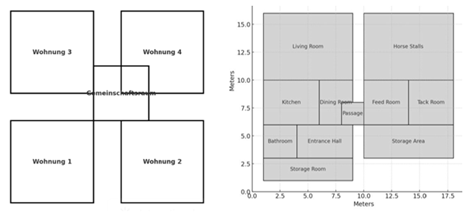}
\vspace{-3pt}
\caption{AI provides 2D floor plans lacking details and violating basic physical laws.}\label{fig:floorplans}
\vspace{-6pt}
\end{figure}

\paragraph{Practicality of design visualization}
It is evident for all interview partners that AI cannot yet be utilized effectively beyond ideation for more detailed planning, as it lacks the capability to display 2D floor plans or 3D models. IP6 criticized the lack of detail that are needed for a real 2D floor plan as well as obvious errors leading to “unrealizable” designs (Figure~\ref{fig:floorplans}).

\subsection{Communication and collaboration with clients}

Interview partners mostly agreed that AI can be integrated into the first customer meeting to establish a basis for design preferences, as it allows one to roughly visualize what the final design might look like though IP4 claims that the usage of AI for the initial meeting may not appear professional and could give an inaccurate impression. IP6 would be cautious to use AI to create designs in real-time, as AI might produce designs that the architect does not like for objective or subjective reasons but the client does. However, there would be great value, if AI could perform simple well defined edits rapidly such as changing building materials (IP6). IP1 asserts that architects must take control of the design process and independently create the designs without relying on an AI tool. IP3 adds that there is a specific target group for whom AI tools may be particularly applicable as AI can serve as an educational tool for beginners, but it has limited application for advanced practitioners.

\begin{figure}
    \centering
    \includegraphics[width=0.5\linewidth]{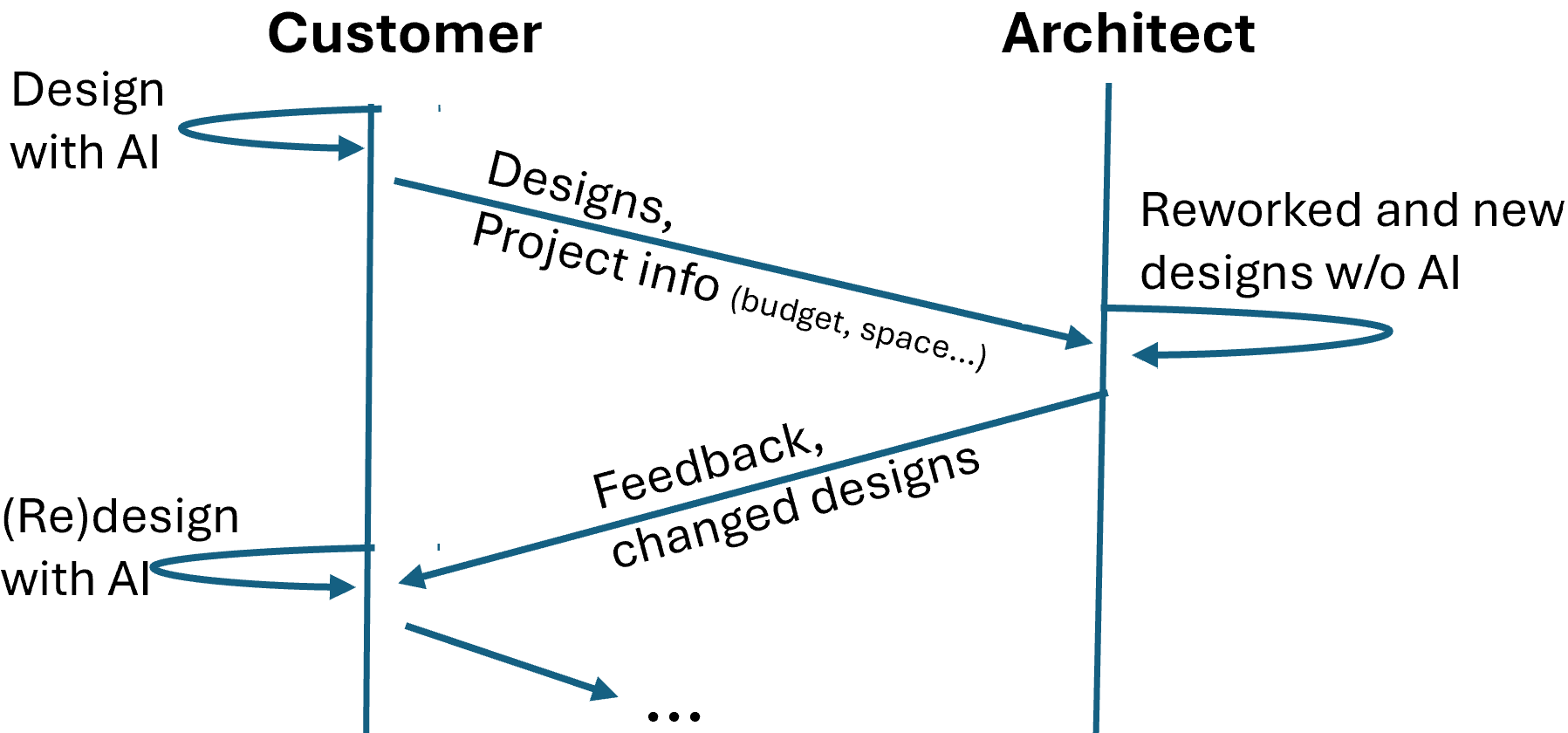}
    \vspace{-3pt}
    \caption{Communication and collaboration with clients (with AI): AI might support architects and clients though interview partners}
    \label{fig:clientInter}
\vspace{-6pt}
\end{figure}

IP3 highlights the benefits of using AI in client discussions, particularly for initial consultations to build consensus on the potential direction of the design. IP1 and IP5 argue that using AI can be very helpful when quick results are needed for clients who have time constraints. IP2 and IP5 caution against the use of AI-generated images within the first client presentation, as they often lack the necessary precision as architects typically work with specific requirements such as particular plots, properties, and building regulations, which AI-generated images may not accurately reflect. IP1 also underscores the significance of sunlight orientation in architectural design as architects pay close attention to where the sun rises and sets, to optimize and maximize natural light within the building, stating that AI is inadequate in this regard, as it struggles to correctly specify window placement, such as placing larger windows on the east side and smaller ones on the north side, due to its lack of knowledge on sunlight orientation and its inability to accurately depict the building’s orientation.

Furthermore, IP1 highlights the critical task of balancing client expectations with practical design solutions as architects can tailor their approach to meet specific client criteria, delving deeply into understanding the true desires of customers. However, this skill involves interpreting the client's often vague vision and translating it into a feasible design and is developed over time as achieving a balance between the client's aspirations and practical, aesthetically pleasing, and structurally sound solutions requires significant expertise. It necessitates a comprehensive approach that considers both large-scale concepts and minute details, ensuring that the overall vision is maintained. Furthermore, according to IP5, architects are key decision-makers who bear responsibility for the outcomes of their projects. Therefore, collaboration with various specialists, such as structural engineers, contractors, and building physicists, is essential. AI is currently incapable of effectively managing these collaborations or replacing the expertise provided by these professionals. The interaction is illustrated in Figure \ref{fig:clientInter}.

\subsection{AI for Feedback}
All interview participants indicated that AI can serve as a feedback source and provide effective input on what has been designed well and what may need enhancement. The utilization of AI aids in refining initial concepts and facilitates the iterative design process by offering constructive critiques and suggestions. IP4 values that AI can highlight both positive and negative aspects. IP1 and IP3 state that the feedback received from AI is very good and actionable. IP2 argues that certain design elements may not require correction because experienced architects might intentionally design these elements in a specific manner. When architects deviate from conventional guidelines and push the boundaries of creativity, AI tends to identify these innovative approaches as problematic and attempts to address them with standardized solutions. This underscores a limitation in AI's ability to recognize and appreciate creative design intentions and misinterpreting them as a design problem, which requires architects to balance AI feedback with personal design choices. However, according to IP1 considering a second opinion is always useful. IP2 addresses the potential disruption of the design process by AI critique as any critique, including that from AI, can potentially derail an architect's concept and lead them in the wrong direction. However, it is stated that the design process inherently involves drawing, erasing, and redrawing elements, a cycle that exists with or without AI and that the introduction of AI will not fundamentally change this iterative nature of the design process. Instead, it may add another layer of feedback that must be carefully evaluated to ensure it aligns with the architect's vision and intentions. Furthermore, IP1 raises concerns about whether AI's critique is genuinely identifying issues or also partially merely echoing the user's own input. As a user, IP1 noted that AI often highlights the same points previously identified as problematic or needing change. For example, when feedback was requested on a house design created within AI, it criticized the very aspects that IP1 as a user intended to alter, such as the addition of balconies to a building. In regards to feedback on design visualization, IP5 adds that AI is advanced enough to provide feedback on floor plans and renderings. It can critique important points, address room layouts, and suggest better arrangements for objects or spaces. Although AI cannot generate floor plans, it can recognize and evaluate them effectively and it can also identify rooms based on the furniture within the image without needing labels and can provide useful critiques, such as indicating if a room is too cramped or if the lighting is insufficient. According to IP5, AI excels in recognizing details within renderings. The primary issue is that it sometimes fails to accurately identify materials, and consequently, it may provide improvement suggestions based on this misinterpretation.

\subsection{Change of Architects’ Role as Creator}
IP1 and IP3 discuss a shift in client interaction and reduced architect involvement as client-driven design minimizes an architect's creative input. Clients may start using AI tools at home to generate plans and only seek architects for verification, which reduces the direct involvement of architects in the initial design process making an architect's role less engaging and more about validating the feasibility of AI-generated designs. IP3 highlights that while clients already bring reference images to architects, the shift towards clients designing final products using AI tools interferes with the traditional role of the architect of actively contributing creatively to the design, reducing the architect's influence on the design. According to IP1, the integration of AI into architecture could significantly alter the role of architects, reducing them to mere executors of client-generated designs as architects might find themselves simply converting client provided images into actionable plans, rather than contributing to the creation of innovative designs. IP6 also mentions that clients already come with ideas in the form of sketches and pictures, but acknowledges that given AI further improves, the role of the architect might shift towards selecting AI-generated designs that are refined and its relevance might reduce. Furthermore, IP3 also touches on the need to distinguish AI-generated images from real reference photos. Clients bringing AI-generated images present a challenge, as these images may not be based on real, feasible projects. Therefore, IP3 further highlights the challenges in managing client expectations and feasibility, noting that architects will increasingly need to navigate client demands while critically assessing the practicality of AI-generated designs. IP5 states that in such scenarios, the architect must first examine and analyze the image, and possibly consult with authorities to determine its feasibility. IP5 also stresses that while AI can create highly complex and innovative designs, many of these ideas are not practically viable due to structural constraints. Similarly, IP6 critizes the lack of knowledge on regulation even if explicitly prompted to stick to them. This highlights the dual nature of architectural work, which, while inherently creative, is also bound by numerous building regulations and rules, creating a contradictory yet essential framework. IP1 highlights a significant concern: If AI were to fully take over the creative processes, the role of architects would become exceedingly dull as their work would be reduced to merely reviewing and approving AI-generated designs, thus stifling their ability to express creativity. Social perception and respect for the architectural profession. With a potential increasing utilization of AI within the architectural field, all interview partners suspect a decrease in social perception and respect for the architectural profession.

\section{Discussion}

\subsection{GenAI Impact on Client–Professional Interactions and Roles in Creative Design}
In “The Death of the Author”, Roland Barthes attests the author little importance (namely none) for his work, since the meaning of the work is generated by the reader. The “author” is a modern figure, created by our society by discovering the “glory of the individual” \cite{r34}. AI-generated designs are once again questioning the extent to which creatives (and non-professionals) are involved in (architectural) designs. Our study aligns with prior works highlighting the change of creative professions. Our work goes beyond the profession but extends towards the design process in a more holistic fashion including also commonly overlooked aspects such as generating feedback. We found interestingly that general-purpose such as Dall-E 3 AI might not be able to generate but still critique well (e.g., 2d floor plans). However, using GenAI systems tailored to such tasks might overcome some of these issues, but nevertheless it is questionable whether such tool become easily accessible for both architects and their clients. The fact that generating a solution is harder than verifying is also deeply embedded in computing theory. Solutions for NP-complete problems can be verified quickly (in polynomial time) but are believed to require much longer to be generated (exponential time).

Our process showing the transformation of GenAI of the design process in Figure~\ref{fig:process} illustrates key findings:
i) AI empowers non-professionals to create visualizations (e.g.\ clients);
ii) AI automates and accelerates process steps (such as sketching);
iii) Creative professions shift towards correcting AI shortcomings (such as lack of context, infeasible designs);
iv) AI can provide constructive feedback that AI designers might use to improve the design.
However, as AI is confined to its training data, it has also been noted that AI might (overly) criticize creative approaches. Also the creativity of AI has been put into question as merely randomly assembling elements, i.e., GenAI models show novelty by recombining the known but not necessarily being of value \cite{r15}. This aligns with a similar claim by \cite{r8}. On the other hand, there are a number of studies showing that human AI collaboration can lead to enhanced creativity in numerous different ways such as by \cite{r17}. GenAI might help laymen and novices in creative professions more than experts — aligning with a finding for call centers employees \cite{r35}. In the light of our study, we conclude that AI is currently more of a supporting tool for human creativity empowering primarily novices but if used on its own without human supervision has severe shortcomings — in generating valuable solutions and appreciating disruptively innovative designs. As such, when viewed from the broader augmentation vs.\ automation perspective \cite{r11}, AI does not (fully) automate the design process. But as AI enables increasingly reasonable design, the initial ideation could be more and more performed by clients with the support of AI replacing or reducing the work for architects for this step. Thus, by focusing on the shifting role boundaries between architects and clients, our study highlights how GenAI reshapes both the production of design artifacts and the negotiation of expertise and authorship. With respect to the creativity process literature \cite{r12} our work highlights that AI as a technology can speed up creative processes though it also underlines the importance of humans to scrutinize AI artifacts for feasibility and contextual fit.

The implications of these changes are profound for architects and their clients. On the one hand, it yields efficiency gains for clients, i.e., AI provides design in seconds rather than hours (with possible loss in revenue for architects), on the other hand, architects might be confronted with impossible to realize AI generations and a lack of reputation and creative requirements of their profession. As such, with the design process AI shows strengths in aesthetics while lacking in structural/civil engineering skills and the failure to comply with building regulations. Furthermore, efficiency gains are not universal. That is, current AI models might be too slow for some tasks such as sketching and the process to articulate and describe a detailed sketch might be slower than the actual sketching.

Architects mention several challenges when using AI, which might stem from both limited AI literacy and technological issues, such as prompt sensitivity \cite{r16} and difficulties in designing prompts \cite{r36}. Also in incremental design process, where only some elements of a design are changed was seen as difficult. That is, as of now, rework to fix AI shortcomings might offset initial efficiency gains.

Furthermore, architects fear that if they use AI some clients might react adversely — a problem not concealed to architecture \cite{r7}. Also if used in face-to-face meeting for interactive designs, architects expressed the fear that AI designs might be appealing for the client but not for them either for subjective or objective reasons. The fact that AI might suggest conflicting ideas those of a human could posit an adoption barrier to AI when it comes to client interactions in addition to prevalent and well-known barriers such as lack of trust, fueling the field of XAI \cite{r37}, and intellectual property and privacy risks \cite{r38}. This calls for architectural “style” alignment as a model requirement for adoption. That is, alignment might go beyond the model level \cite{r39} but to the level of an individual in areas where aspects such as personal style and expression are essential.

While AI is still seen as valuable for ideation, surprisingly, the more mechanistic, and more standardized translation of early ideation into detailed plans, turns out to be difficult to automate through AI and remains as the most lucrative part of the planning process with architects — for now. As a limitation, we focused on just one tool and a small set of professional architects mostly from the same digital era, though technological shortcomings have been pointed out also for other text-to-image tools \cite{r8}. Furthermore, as AI is rapidly evolving, some of AI’s shortcomings might be less relevant in the future.

\subsection{Broader Societal Reflections on Dynamics of Non-experts, Experts and AI}
GenAI also triggers profound societal and philosophical changes in the triangle AI vs.\ experts vs.\ non-experts (most of society).
\paragraph{Shifts in Layperson–Expert Interactions: Knowledge and Skill Dynamics.}
Historically, access to specialized knowledge and skills has defined power dynamics between experts and laypersons, i.e., most of society. Before widespread digitalization, specialized knowledge resided primarily within physical libraries and expert communities, making it labor-intensive or even inaccessible for laypersons. This led to considerable information asymmetry and reinforced experts' authority, power, economic well-being, and societal status.

The advent of search engines, notably Google, represented the first significant reduction of information asymmetry. Clients increasingly arrived at expert consultations with preliminary knowledge obtained online, altering traditional dynamics:
\begin{itemize}
\item Individuals can approach architects with inspiration images sourced from Google or Instagram.
\item Legal clients can enter consultations with information about applicable laws and similar cases.
\item Medical patients can present self-diagnoses based on symptom searches.
\end{itemize}
Yet, despite increased access, experts largely retained authority due to multiple reasons: 1) the complexity of interpreting specialized information (e.g., legal cases documented in legal language by courts, scientific articles using scientific formalism and expressions); 2) the effort to identify relevant information, i.e., search engine rank documents based on their relevance. However, these documents must be read and filtered by a human;  3) the effort of combining and reasoning about this information, i.e., even given all relevant information for a particular problem, a user might have to draw non-trivial conclusions to address her specific task. For example, if a user wants to know the maximum height of a house, he can build on a piece of 500m\textsuperscript{2} of land, the search engine might return that the maximum height is “square root of size of land – distance in meter to border of land”, which needs further reasoning.

GenAI addresses at least partially the above problems of classical search, by shifting not only knowledge but also applied skills directly into laypersons' hands. Now, clients may:
\begin{itemize}
\item Visit an architect with fully visualized house designs generated by AI.
\item Consult lawyers with AI-generated legal strategies tailored for specific situations.
\item Engage healthcare professionals with AI-assisted tailored self-diagnoses and treatment plans.
\end{itemize}

\begin{figure}[ht]
\vspace{-6pt}
\centering
\includegraphics[width=0.7\linewidth]{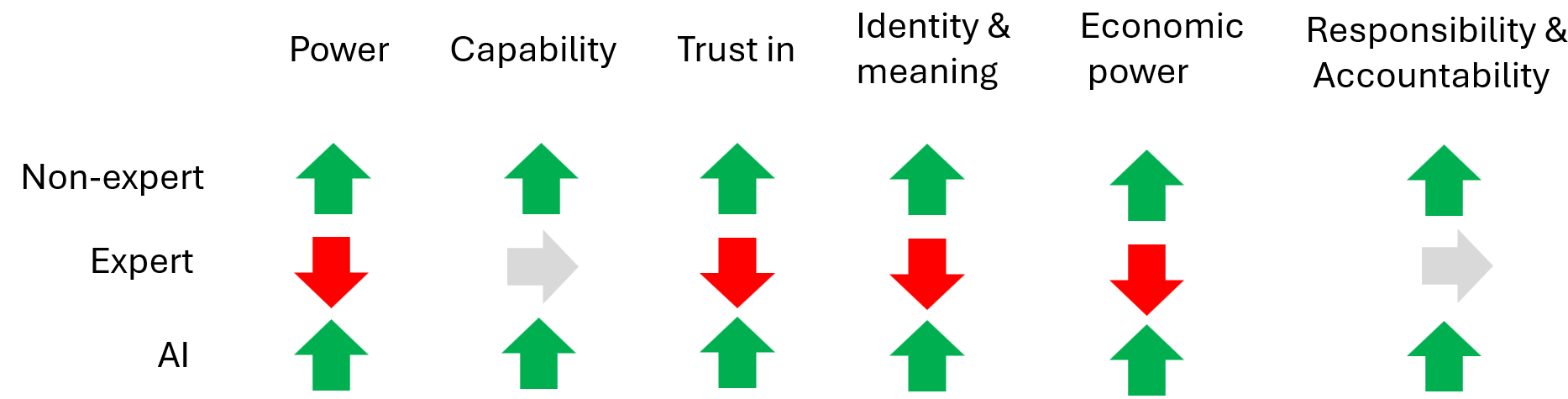}
\caption{Possible changes in key dimensions for non-experts, experts, and (Gen)AI.}\label{fig:triangle}
\vspace{-10pt}
\end{figure}

\paragraph{The Triangle of Clients, Experts, and AI: Who Gains and Who Losses?}
We discuss the dimensions shown in Figure~\ref{fig:triangle} from the three actors non-expert, expert and AI. With respect to power, i.e., the ability to influence decisions, assert control, and shape outcomes, non-experts benefit as they are empowered through knowledge by the AI. In turn, also the power of AI raises as more people rely and follow its advice. From the perspective of a non-expert engaging with a human expert, AI might be viewed as a second expert with which one can engage free of charge at any time. This is in stark contrast to engagements with costly and often unavailable human experts. Thus, the availability of AI significantly reduces barriers to expert-level insights, enhancing a client's perceived power in interactions with human experts, who may increasingly need to justify or defend their recommendations against those provided by AI. Interaction with AI improves a non-expert’s capability as she can learn more about her issues. While experts can also benefit from AI, their benefits in terms of capability is much less. As AI is used more by clients, more data is generated that can be leveraged by AI for learning. While data quality by clients can be worse than those of experts, overall more data is helpful. As clients capabilities increase and increasingly humans get outperformed by AI, the trust in AI and themselves increases, while those in experts diminishes. Such a shift in trust dynamics could accelerate rapidly if clients perceive tangible personal benefits from AI, potentially leading to diminished trust in traditional institutions or professional expertise and heightened confidence in AI solutions. These developments might come at a cost of a reduced identity and meaning for experts, while they increase those of laymen as they become more self-reliant and feel more equal. Consequently, professional roles may become increasingly ambiguous, potentially triggering an identity crisis among experts who must redefine their societal and professional value beyond tasks that can be efficiently handled by AI. Their required skillset might shift: Appropriate and convincing communication might gain in importance, as they can rely less on their authority. The economic power of experts vanishes as, e.g., a lawyer or architect cannot charge as much as AI performs part of their work and also people might be more willing to not use such expert services in the first place. In turn, non-experts benefit from cheaper access to expertise. This shift could also disrupt traditional economic structures, forcing experts to either significantly differentiate their offerings, specialize in AI-resistant niches, or adopt hybrid models blending human and AI services. Thus, as clients feel more empowered and self-sufficient and accessing experts is tedious and costly, they might more and more decisions without consulting experts. This is concerning in areas, where AI-assisted perform worse than experts. For example, assume a parent observes medical symptoms at a child and decides to self-diagnose using AI rather than spending time and money for a doctor.  Thus, the increased power of clients and AI also comes with increased responsibility and accountability. Clients, now more autonomous, must acknowledge their heightened individual responsibility in cases of independent AI use, particularly in sensitive or high-stakes domains such as health, safety, or legal compliance. 


\paragraph{Long term impact on cultural heritage}
GenAI can take the crucial role of suggesting architectural designs based on prompts by architects and laymen. In theory, prompts and, in turn, users control the design, but in practice control is limited because (i) AI does not always follow prompts accurately and (ii) it is difficult to express designs in great detail. In turn, GenAI which can combine learnt design elements in its training data in novel way, will have a profound impact on designs. This raises a few concerns: (a) Less innovations: AI lacks creativity beyond recombination and adaption \cite{r15}, (b) Less diversity: For identical or even just similar prompts diversity of AI is generally more limited than that of humans, i.e., varying outputs of AI are considered an issue known as prompt sensitivity \cite{r16} that should be prevented, (c) Loss of cultural heritage: AI is driven by its training data, which is often dominated by large, English speaking countries. For example, smaller European countries might be less represented and, in turn, might be dominated by design influences of other continents.  Explicit prompting (e.g., design an Italian style house) might partially offset such biases towards specific cultures though this is not clear. Thus, in conclusion, it is critical to carefully monitor ongoing developments and AI tools for biases and usage, and potentially shape future versions of AI tools to ensure culture is shaped in a desired manner.

\section{Conclusions}
Our study shows that Generative AI might soon transform the architectural design process, particularly by empowering clients to engage in ideation and visualization pushing the role of the architects more towards a curator. Still, also architects might reap some benefits of obtaining design ideas and feedback on their design. But as of now, AI struggles with adhering to basic structural engineering principles, regulations, and the specific building context. On a broader scale, we elaborated on the shift that GenAI might bring among a number of actors and more broadly on our cultural heritage.




\end{document}